\documentclass{article}

\PassOptionsToPackage{numbers, compress}{natbib}
\usepackage[preprint]{neurips_2021}

\usepackage[utf8]{inputenc} %
\usepackage[T1]{fontenc}    %
\usepackage[hidelinks]{hyperref}
\usepackage{url}            %
\usepackage{booktabs}       %
\usepackage{amsfonts}       %
\usepackage{nicefrac}       %
\usepackage{microtype}      %
\usepackage{xcolor}         %
\usepackage{colortbl}
\usepackage{graphicx}       %
\usepackage{setspace}
\usepackage{enumitem}
\usepackage{multirow}
\usepackage{caption}
\usepackage{dirtytalk}
\usepackage{mathtools}
\usepackage{xspace}
\usepackage{soul}
\usepackage{amssymb}
\usepackage{pifont}
\newcommand{\cmark}{\ding{51}}%
\newcommand{\xmark}{\ding{55}}%
\graphicspath{{fig/}}

\usepackage[disable]{todonotes}
\definecolor{darkgreen}{rgb}{0,0.45,0}

\newcommand{\dq}[1]{\todo[fancyline,color=orange!50]{\textbf{Dequan}: #1}\ignorespaces}
\newcommand{\es}[1]{\todo[fancyline,color=green!50]{\textbf{Evan}: #1}\ignorespaces}
\newcommand{\se}[1]{\todo[fancyline,color=yellow!50]{\textbf{Sayna}: #1}\ignorespaces}

\newcommand{\method}{$\text{method}$\xspace}

\makeatletter
\newcommand*{\etc}{%
    \@ifnextchar{.}%
        {etc}%
        {etc.\@\xspace}%
}
\makeatother

\DeclarePairedDelimiterX{\infdivx}[2]{(}{)}{%
  #1\;\delimsize|\delimsize|\;#2%
}
\newcommand{\kld}[2]{\ensuremath{D_{KL}\infdivx{#1}{#2}}\xspace}

\title{On-target Adaptation}

\author{%
Dequan Wang \\
UC Berkeley \\
\And
Shaoteng Liu \\
CUHK \\
\And
Sayna Ebrahimi \\
UC Berkeley \\
\AND
Evan Shelhamer \\
Imaginary Number \\
\And
Trevor Darrell \\
UC Berkeley \\
}

\begin{document}

\maketitle

\begin{abstract}
    Domain adaptation seeks to mitigate the shift between training on the \emph{source} domain and testing on the \emph{target} domain.
    Most adaptation methods rely on the source data by joint optimization over source data and target data.
    Source-free methods replace the source data with a source model by fine-tuning it on target.
    Either way, the majority of the parameter updates for the model representation and the classifier are derived from the source, and not the target.  %
    However, target accuracy is the goal, and so we argue for optimizing as much as possible on the target data.
    We show significant improvement by on-target adaptation, which learns the representation purely from target data while taking only the source predictions for supervision.
    In the long-tailed classification setting, we show further improvement by on-target class distribution learning, which learns the (im)balance of classes from target data.
\end{abstract}

\section{Introduction}

Deep networks achieve tremendous success for various visual tasks at the expense of massive data collection and annotation efforts.
Even more data is needed when training (source) and testing (target) data do not come from the same distribution, as the model needs to be fine-tuned on the new data in a supervised or unsupervised way.
To reduce the annotation burden when dealing with new data, unsupervised domain adaptation (UDA) approaches transfer knowledge from source to target without annotating target data.
In order to do so, UDA requires simultaneous access to the annotated source and unannotated target data.
However, this requirement may not be entirely practical, in that \emph{shifted} or \emph{future} target data may not be available during training.
Furthermore, (re-)processing source data during testing may be limited by computation, bandwidth, and privacy.
Rather than retain the source data, we turn to focus more on the target data.

Current frontiers in adaptation seek to adapt without source data or even do so during testing.
Source-free domain adaptation transfers knowledge from the source model parameters to the target data without joint use of the source data~\cite{liang2020source,li2020model,kundu2020universal}.
Test-time adaptation updates source parameters on-the-fly as target data is received during testing~\cite{sun2019test, schneider2020improving, wang2021tent}.
The general outline of test-time adaptation methods is to first choose an unsupervised objective---such as entropy minimization, rotation prediction, and the like---and then gradually update by gradient optimization of this unsupervised loss over the target data.
While these test-time methods update on target data, they initialize from the source parameters alone.
As shown by the accuracy improvements of prior work, this can be a suitable initialization, especially when only the source model and target data are available.
However, since better target accuracy is the final goal, we argue for learning even more from the target data.

We introduce on-target adaptation to separate the target representation from the source representation.
By factorizing the representation from the classification, we can train the representation entirely on the target data by self-supervision.
Given this on-target representation, we can then distill the source predictions into a new classifier, without the architectural constraints needed to transfer the source representation.
We make effective use of test-time adaptation, contrastive learning, and teacher-student transfer learning to do so with state-of-the-art accuracy and computational efficiency on common domain adaptation benchmarks.

We turn test-time adaptation methods into teacher models, generating pseudo-labels with the adapted source model.
Then we initialize the target model with self-supervised learning methods.
The goal of contrastive learning is to learn a generalizable visual representation from large-scale unlabeled images.
The recent advances~\cite{chen2020exploring, caron2020unsupervised, he2020momentum, chen2020improved, grill2020bootstrap, chen2020exploring, zbontar2021barlow} demonstrate that self-supervised models are equipped with competitive representation compared to the supervised features.
In our work, we leverage the most recent four off-the-shelf methods (MoCo v2~\cite{chen2020improved}, SwAV~\cite{caron2020unsupervised}, SimSiam~\cite{chen2020exploring}, Barlow Twins~\cite{zbontar2021barlow}) to initialize on-target representation.

In this way, our target feature could be initialized in a totally target-specific way, while at the same time delivering the freedom to choose a more computationally efficient architecture.
Typically, the source model has more parameters, in order to better fit the source data with this capacity.
The target data, being shifted, may not need this full capacity, so we can choose a smaller network for adaptation.
In doing so we significantly reduce the computational cost of inference and learning for the target domain.
Finally, we connect the on-target representation with the source task by transferring source predictions via teacher-student learning.
Specifically, we transfer predictions from the supervised source model (teacher) to the target model with its unsupervised representation (student) via semi-supervised learning using consistency regularization and pseudo-labeling.

Despite its simplicity, the empirical results on both domain adaptation and long-tailed recognition benchmark datasets demonstrate the proposed \method outperforms the state-of-the-art methods given only the unlabeled target data.
For example, our \method brings around 3\% absolute improvement compared to state-of-the-art unsupervised and source-free domain adaptation methods on various benchmarks, such as VisDA-C~\cite{peng2017visda}, ImageNet Sketch~\cite{wang2019learning}, Office Home~\cite{venkateswara2017deep}, while reducing 50\%+ parameters and runtime flops, and 75\%+ memory consumption at each feed-forward pass on the target model.
Analysis experiments echo our statement of on-target feature learning, check sensitivity to network architecture/initialization, contrastive learning method, loss function, amount of optimization, and back the generality over architectures and initializations.

\textbf{Our contributions}
\begin{itemize}[leftmargin=7mm, topsep=0.5mm, itemsep=0.0mm]
  \item We decouple the representation and classifier to fit the target data better.
    In doing so, we demonstrate state-of-the-art adaptation by contrastive learning and teacher-student transfer learning.
    These results are due to our insight that we can transfer source predictions without the source representation.
  \item We distill source predictions into distinct target architectures.
    By not inheriting the source representation, we can adapt more computationally-efficient architectures without sacrificing accuracy.
    In fact, with teacher-student learning, we can even improve adaptation accuracy with a smaller student model.
  \item Experiments on both domain adaptation and long-tailed recognition benchmarks demonstrate that our on-target framework outperforms the state-of-the-art methods without access to the source data or any change of the source-domain training procedure.
\end{itemize}

\section{Related work}

\textbf{Adaptation}
On-target adaptation is unique in its decoupling of the target representation from the source representation.
Prior adaptation approaches transfer the source representation to the target, either by joint optimization or by initialization.
To transfer the source model to a visually different target domain, unsupervised domain adaptation (UDA) learns a joint representation for both domains for visual recognition tasks, such as image classification~\cite{tzeng2014deep}, object detection~\cite{chen2018domain}, semantic segmentation~\cite{hoffman2016fcns}.
Some of the most representative unsupervised domain adaptation ideas are 1) maximum mean discrepancy~\cite{long2015learning,long2017deep}; 2) moment/correlation matching~\cite{sun2016return,zellinger2017central}; 3) domain confusion~\cite{ganin2015unsupervised,tzeng2017adversarial}; 4) GAN-based alignment~\cite{liu2017unsupervised,hoffman2018cycada}.
All these UDA methods need simultaneous access to both source and target data.
In practice, it might be impossible to meet this requirement due to the limited bandwidth, computational power, or privacy concerns.
Therefore, some recent works focus on test-time training~\cite{sun2019test}, source-free adaptation~\cite{liang2020we}, and fully test-time adaptation~\cite{wang2021tent} settings,
in which a given source model is fine-tuned on the target data without source data.

\textbf{Semi-supervised learning}
Many UDA methods follow the practice of semi-supervised learning, especially pseudo labeling~\cite{lee2013pseudo} which is to utilize the model prediction to generate supervision for the unlabeled images.
The typical setup of unsupervised domain adaptation methods is to jointly optimize with ground truths on the source and pseudo labels on the target~\cite{zhang2018collaborative,choi2019pseudo,long2017deep,zou2018unsupervised}.
When source data annotations are not available, DeepCluster~\cite{caron2018deep} and SHOT~\cite{liang2020we,liang2020source} further leverage weighted k-means clustering to reduce the side effects on noisy pseudo labels.
Similarly, our method does not require access to labeled source data, 
while only relying on the target images with generated pseudo labels. %
In addition, our method heavily benefits from the contrastive learned target domain representation,
which is treated as initialization to overcome the misleading of noisy pseudo labels.

\textbf{Long-tailed recognition}
The recent studies on long-tailed recognition, which is proposed to tackle the imbalanced distribution in real-world data, could be divided into three groups: 1) re-balancing data distribution~\cite{chawla2002smote,han2005borderline,shen2016relay,mahajan2018exploring}; 2) class-balanced loss design~\cite{cui2019class,khan2017cost,cao2019learning,khan2019striking,huang2019deep,lin2017focal,shu2019meta,ren2018learning,hayat2019max}; 3) transfer learning from head to tail~\cite{yin2019feature,liu2019large}.
All these related works focus on tuning the training procedure to overcome the imbalanced sample distribution so that the model could learn class-balanced features and classifier with the given long-tailed distributed training images.
In our work, we argue that we could overcome these challenges merely during the test-time.
To the best of our knowledge, our work is the first to calibrate a classifier for long-tailed recognition fully during testing.

\section{Method: on-target adaptation}
\label{sec:method}

\begin{figure}[t]
	\centering
	\includegraphics[width=0.95\linewidth]{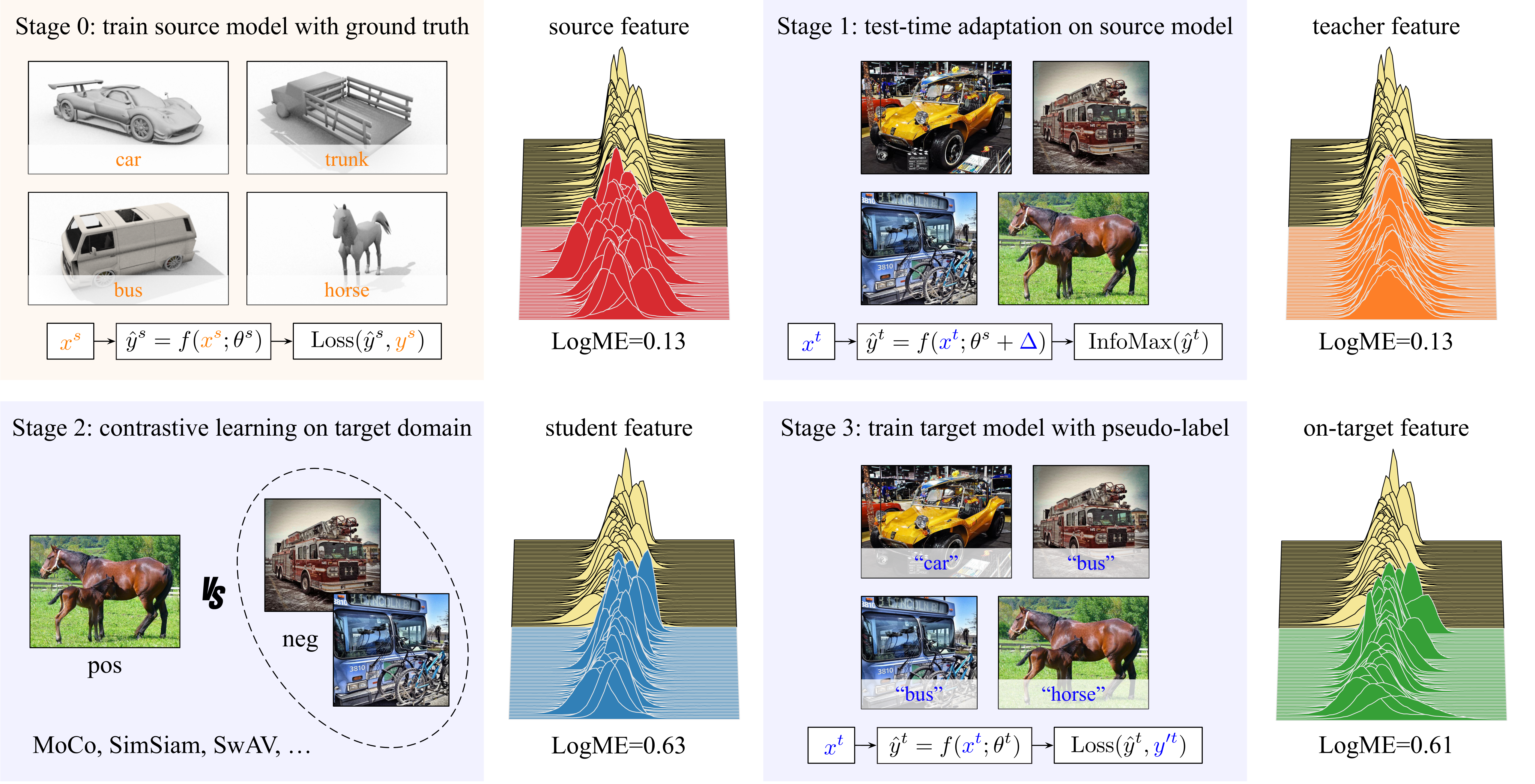}
	\caption{Illustration of our \method at each stage with feature distribution. Orange color at stage 0 shows training-time and blue color at stage 1/2/3 represents test-time operations. We use LogME~\cite{you2021logme} as a practical assessment of features and labels correlation for transfer learning (higher means stronger correlation). We visualize the representation of each stage (front) with the corresponding reference feature (back). Best viewed in color.}
	\label{fig-framework}
	\vspace{-5mm}
\end{figure}

The goal of the proposed on-target adaptation is to tackle domain shift during the test-time with only a source model, without the access of annotation and source data.
Specifically, the supervised model with source parameter $f(\cdot; \theta^s)$ trained on source images $x^s$ and labels $y^s$ needs to generalize on unlabeled target data $x^t$ when an unneglectable domain shift happened.
Our on-target adaptation (Figure~\ref{fig-framework}) is proposed to obtain target model parameter $\theta^t$ purely during test-time.

\textbf{Stage 0 (source): train model with labeled source data}
We train a deep ConvNet and learn source parameter $\theta^s$ by minimizing vanilla cross-entropy loss $\mathcal{L}(\hat{y}^s, y^s)$ on labeled source data $(x^s, y^s)$.
Specifically, $\mathcal{L}(\hat{y}^s, y^s)=-\Sigma_c p(y^s_c)\log(p(\hat{y}^s_c))$ for the predicted probability $\hat{y}^s_c$ of class $c$, where target probability $y^s_{gt}$ is 1 for the ground truth class $gt$ and 0 for the rest.
\textbf{Stage 1 (teacher): adapt source model during test-time}
We update the source parameter $\theta^s$ during testing to minimize information maximization (InfoMax) loss~\cite{gomes2010discriminative}.
Specifically, InfoMax loss augment entropy loss $\mathcal{L}_{ent}=-\Sigma_c p(\hat{y}^t_c)\log(p(\hat{y}^t_c)$ with diversity  objective $\mathcal{L}_{div}=\kld{\hat{y}^t}{\frac{1}{C}\mathbf{1}_{C}} - \log(C)$.
where $D_{KL}$ indicates the Kullback–Leibler divergence, $\mathbf{1}_{C}$ is an all-one vector with $C$ dimensions.
Here $\frac{1}{C}\mathbf{1}_{C}$ indicates the target label vector with evenly distributed $\frac{1}{C}$ probabilities,
where $\mathcal{L}_{div}$ is propose to enforce the global diversity over classes.

As for the parameters to optimize over, we follow the motivation of decoupling the representation and classifier.
When the classifier is frozen, the goal of optimization is to mitigate domain shift by deriving proper target features from the source model.
In particular, we keep the classifier the same on both source and target domain, and obtain $\Delta$ by the gradient of the test-time objective (InfoMax), to update the representation part of model parameter $\theta^s$.

\textbf{Stage 2 (student): initialize target model with contrastive learning}
Instead of fine-tuning from source model, we choose to initialize the target feature purely from target data.
Benefiting from the recent advances in contrastive learning methods, we train an unsupervised model with purely unlabeled target images.
Specifically, we initialized target representation via improved momentum contrast learning (MoCo v2)~\cite{he2020momentum, chen2020improved}.
It is worth noting that our method does not require a specific contrastive learning method.
In other words, the default MoCo v2 could be easily replaced by a more recent self-supervised learning model, such as SwAV~\cite{caron2020unsupervised}, SimSiam~\cite{chen2020exploring}, Barlow Twins~\cite{zbontar2021barlow}.
Such a modular design makes it easier to benefit from the latest advance in contrastive learning. 
We have performed an ablation study on the choice of contrastive learning method in Section~\ref{sec:exp-ablation}.

\textbf{Stage 3 (teacher-student): transfer knowledge from teacher to student}  We utilize the test-time adapted source model $f(\cdot; \theta^s+\Delta)$ as the initial teacher model to generate pseudo labels $y^{\prime t}$ on unannotated target images $x^t$.
Then we fine-tune the student model $f(\cdot; \theta^t)$ initialized by contrastive learning on target data with cross-entropy loss $\mathcal{L}(\hat{y}^t, y^{\prime t})=-\Sigma_c p(y^{\prime t}_c)\log(p(\hat{y}^t_c))$.
The teacher would be replaced with the latest student to gradually denoise pseudo labels for the subsequent phase.
Meanwhile, the contrastive learned model would re-initialize the student feature to eliminate the accumulated errors from imperfect pseudo labels.
In other words, the student model would start over one more time with only newer pseudo labels for the next transferring phase.
\dq{We need to highlight that contrastive feature has been used every cycle!}

\begin{figure}[t]
	\centering
	\includegraphics[width=0.95\linewidth]{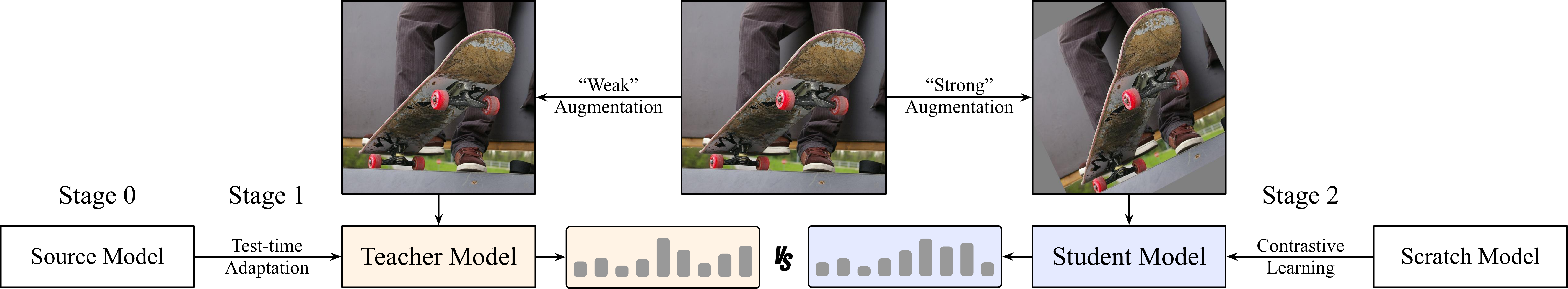}
	\caption{Illustration of teacher-student (stage 3) in our \method. Transfer learning between the teacher (orange) and the student (blue), where pseudo labels are generated on the weakly-augmented images. The model is trained on the strongly-augmented target data to match the pseudo labels.} %
	\label{fig-teacher-student}
	\vspace{-6mm}
\end{figure}

Figure~\ref{fig-teacher-student} illustrates the procedure of transferring the knowledge from teacher to student. Specifically, the interaction between teacher and student models benefits from consistency regularization and pseudo-labeling, inspired by a recent semi-supervised learning approach called FixMatch~\cite{sohn2020fixmatch}.
During the transferring, we augment the target images with random cropping, random flipping, and AutoAugment with ImageNet policy, as \say{strong} augmentation, while the \say{weak} augmentation is the combination of resizing and center cropping when generating pseudo labels.
Relying on the assumption that the model should generate similar predictions on data-augmented versions of the same image~\cite{bachman2014learning, sajjadi2016regularization, laine2016temporal}, consistency regularization enforces the cross-entropy loss between student output on strongly-augmented images and teacher output on weakly-augmented images.

\section{Experiments}
\label{sec:experiments}

\subsection{Setup}
\label{sec:exp-setup}

\textbf{Datasets}
We intensively evaluate our method on both domain adaptation and long-tailed recognition benchmarks, including VisDA-C~\cite{peng2017visda}, Office Home~\cite{venkateswara2017deep}, Sketch~\cite{wang2019learning}, ImageNet-LT~\cite{liu2019large}, and iNaturalist-18~\cite{van2018inaturalist}. Figure~\ref{fig-domain-shift} presents some of example images to illustrate domain shifts.

\textbf{Metric}
We report top-1 accuracy (denoted as acc.) on the whole dataset for all datasets.
As for VisDA-C, we additionally report the percentage accuracy of each category and the corresponding average of all categorical accuracies (denoted as avg.), due to the imbalanced distributed label space.
As for Office Home, we calculate the average for all kinds of domain shifts as the summary number for each method.
As for long-tailed recognition benchmarks, %
we additionally report percentage accuracy on many-shot (more than 100 samples), medium-shot (20-100 samples), and few-shot (less than 20 samples) following the evaluation protocol from \cite{liu2019large,kang2019decoupling}.

\textbf{Baselines}
We choose the most recent fully test-time adaptation method, TENT~\cite{wang2021tent}, and source-free adaptation framework, SHOT~\cite{liang2020we}, as our adaptation baselines.
Entropy minimization is the target optimization objective for both TENT and SHOT.
SHOT additionally regularizes optimization by the information maximization (InfoMax) loss~\cite{gomes2010discriminative,shi2012information,hu2017learning} to augment entropy minimization on each sample with diversity maximization across samples.
Alongside their role as baselines, these methods can serve as teacher models for our stage 1.
We also compare with unsupervised domain adaptation (UDA) baselines, including DANN~\cite{ganin2015unsupervised}, DAN~\cite{long2015learning}, ADR~\cite{saito2018adversarial}, CDAN+E~\cite{long2018conditional}, CDAN+BSP~\cite{chen2019transferability}, CDAN+TN~\cite{wang2019transferable}, SAFN~\cite{xu2019larger}, SWD~\cite{lee2019sliced}, DSBN+MSTN~\cite{chang2019domain}, STAR~\cite{lu2020stochastic}.
It is worth noting that all these UDA methods are fine-tuning from the ImageNet pretrained ResNet-101 source model, with access to both source and target data.
TENT and SHOT likewise initialize their representation from the source model.
In contrast, our method trains ResNet-18 from scratch, and entirely on the target data, by contrastive learning for the representation and teacher-student learning for the classification.

As for long-tailed recognition, we choose learnable weight scaling (LWS)~\cite{kang2019decoupling} as the train-time baseline.
LWS first decouples the full model into representation and classification, and then only adjusts the magnitude of the classifier with class-balanced sampling.
Our on-target class distribution learning is a test-time extension of LWS, which re-scales the classifier on unlabeled target data.

\textbf{Architecture}
For comparability with state-of-the-art models, we choose 18/50/101-layer ResNet models~\cite{he2016deep} for both main results and ablation studies. When reproducing the prior works, we keep the architecture the same, for example, the weight-normalization~\cite{salimans2016weight} augmented ImageNet pretrained ResNet-101 for SHOT~\cite{liang2020we}.

\subsection{Implementation}
\label{sec:exp-implementation}

Our implementation is in PyTorch~\cite{paszke2019pytorch} and depends on VISSL~\cite{goyal2021vissl}, MMClassification~\cite{2020mmclassification}, Weights \& Biases~\cite{wandb} libraries.
The code will be made publicly released for publication.
\dq{we need to name all these stages and the corresponding combo (such as stage 0+1+3)}

\textbf{Stage 0 (source)}
We train residual networks~\cite{he2016deep} with various depths (including 18, 50, 101), and initializations (ImageNet pretrain or Kaiming init~\cite{he2015delving} when training from scratch).
We optimize cross-entropy loss by SGD with an initial learning rate 0.1, momentum 0.9, weight decay 0.0001, batch size 256.
Label smoothing technique~\cite{muller2019does} could be leveraged to further improve the discriminability of the source model if it is needed by the next stage.%
 We follow the engineering of data augmentation from ImageNet training practice, such as random cropping, random flipping, and color jitter. We choose ImageNet statistics as the default input mean and variance for all models.

\textbf{Stage 1 (teacher)}
We experiment with three test-time models as teachers: 1) source-only, 2) TENT~\cite{wang2021tent}, 3) SHOT~\cite{liang2020we}.
As for TENT, We replace the original entropy with SHOT's InfoMax loss.
\es{Tent and SHOT require more explanation, as well as contrast, because right now it's quite unclear what these methods are. The baseline paragraph could briefly review entropy minimization, and then identify their adaptation parameters.}
To accompany such a modification, we choose SGD with a learning rate 0.0001, momentum 0.9, weight decay 0.0001, different from the original Adam~\cite{kingma2015adam} optimizer.
In addition to batch normalization~\cite{ioffe2015batch}, we also update convolutional layers except the final classification layer.
As for SHOT, We execute the authors' open-sourced codebase with the same hyper-parameters for various architectures (ResNet-50 and ResNet-101), initialization (from scratch and ImageNet pretrain), and domain shifts (train to val/test splits on VisDA-C).

\textbf{Stage 2 (student)}
We experiment with two designs as students: 1) source-only, 2) contrastive learning.
Specifically, we leverage some of off-the-shelf contrastive learning methods to initialize target-domain representation, such as MoCo v2~\cite{chen2020improved}, SimSiam~\cite{chen2020exploring}, SwAV~\cite{caron2020unsupervised}, Barlow Twins~\cite{zbontar2021barlow}.
Compared to their training recipes on ImageNet, we have more epochs on VisDA-C val/test with the same batch size, learning rate, data augmentation, and model architecture, to make the training procedure longer with the smaller amount of images.

\textbf{Stage 3 (teacher-student)}
By default, the whole knowledge distillation consists of three phases, where each phase has 10 epochs to train the student model with the hard pseudo label.
The student would be reset to the contrastive model to avoid error accumulation at the beginning of every phase.
The teacher would be replaced with the latest student before starting the next phase,
so that the quality of pseudo-labeling could be improved gradually.
We utilize SGD with an initial learning rate 0.01, momentum 0.9, weight decay 0.0001, batch size 256, and cosine annealing scheduler~\cite{loshchilov2016sgdr}.

\subsection{On-target adaptation}
\label{sec:exp-adaptation}
\begin{table}[t]
\centering
\Huge
\resizebox{0.95\textwidth}{!}{
    \begin{tabular}{lccccccc}
\multirow{2}{*}{method} & \#1 test-time & \#2 contrastive & \#3 teacher & \multicolumn{2}{c}{VisDA-C train} & Imagenet & Office \\ %
    & adaptation & learning & student & $\rightarrow$val & $\rightarrow$test & $\rightarrow$Sketch & Home \\ %
\midrule
source-only & ~ & ~ & ~ & 21.8 & 23.9 & 27.6 & 51.7 \\ %
test-time adaptation & \cmark & ~ & ~ & 31.4 & 34.3 & 35.6 & 53.8 \\ %
\midrule
on-target adaptation & ~ & ~ & \cmark & 28.3 & 31.8 & 27.9 & 54.3 \\ %
without contrastive & \cmark & ~ & \cmark & 43.3 & 46.0 & 40.5 & 56.8 \\ %
\midrule
\multirow{2}{*}{on-target adaptation} & ~ & \cmark & \cmark & 29.1 & 33.9 & \xmark & \xmark \\ %
    & \cmark & \cmark & \cmark & 49.9 & 51.2 & \xmark & \xmark \\ %
\bottomrule 
\end{tabular}}
\vspace{2mm}
\caption{Classification accuracy of our \method at each stage on domain adaptation benchmarks including VisDA-C~\cite{peng2017visda}, ImageNet Sketch~\cite{wang2019learning}, and Office Home~\cite{venkateswara2017deep}. We use ResNet-18 as target model. We initialize ResNet-50 source model from scratch/ImageNet pretrained for VisDA-C/others. 
}
\label{tab-stage-summary}
\vspace{-5mm}
\end{table}

Table~\ref{tab-stage-summary} reports numbers for the individual contribution for each stage on domain adaptation datasets.
Starting with the source-only model as the baseline, the test-time adaptation (stage 1) brings the significant improvement on the top of the source model.
Then, teacher-student (stage 3) leverages the source-only or test-time adapted model as the initial teacher model.
Here we name the model which generates pseudo labels as teacher model, the model which learns from pseudo labels as student model.
The student model (source or contrastive learned model) is trained with the pseudo labels generated by the teacher.
After several phases of transfer learning, the student outperforms the initial teacher by a large margin.
Furthermore, the contrastive learned student (stage 2) boosts the knowledge transferring efficiency with a more representative target domain feature to start with.

\begin{table}[t]
\centering
\resizebox{0.95\textwidth}{!}{
\Huge
\begin{tabular}{lcrrrrrrrrrrrr>{\columncolor[gray]{.9}}r>{\columncolor[gray]{.9}}r}
method & network & plane & bcycl & bus  & car  & horse & knife & mcycl & person & plant & sktbrd & train & trunk & avg. & acc. \\
\midrule
ADR~\cite{saito2018adversarial} & R101P & 94.2  & 48.5  & 84.0 & 72.9 & 90.1  & 74.9  & 92.6  & 72.5   & 80.8  & 61.8   & 82.2  & 28.8  & 73.6 & 73.8 \\
CDAN+E~\cite{long2018conditional} & R101P & 85.2  & 66.9  & 83.0 & 50.8 & 84.2  & 74.9  & 88.1  & 74.5   & 83.4  & 76.0   & 81.9  & 38.0  & 73.9 & 71.0 \\
CDAN+BSP~\cite{chen2019transferability} & R101P & 92.4  & 61.0  & 81.0 & 57.5 & 89.0  & 80.6  & 90.1  & 77.0   & 84.2  & 77.9   & 82.1  & 38.4  & 75.9 & 73.4 \\
SAFN~\cite{xu2019larger} & R101P & 93.6  & 61.3  & 84.1 & 70.6 & 94.1  & 79.0  & 91.8  & 79.6   & 89.9  & 55.6   & 89.0  & 24.4  & 76.1 & 75.6 \\
SWD~\cite{lee2019sliced}  & R101P & 90.8  & 82.5  & 81.7 & 70.5 & 91.7  & 69.5  & 86.3  & 77.5   & 87.4  & 63.6   & 85.6  & 29.2  & 76.4 & 75.6 \\
DSBN+MSTN~\cite{chang2019domain} & R101P & 94.7 & 86.7 & 76.0 & 72.0 & 95.2 & 75.1 & 87.9 & 81.3 & 91.1 & 68.9 & 88.3 & 45.5 & 80.2 & 79.2 \\
STAR~\cite{lu2020stochastic} & R101P & 95.0 & 84.0 & 84.6 & 73.0 & 91.6 & 91.8 & 85.9 & 78.4 & 94.4 & 84.7 & 87.0 & 42.2 & 82.7 & 80.4 \\
\midrule
SHOT~\cite{liang2020we} & R101P & 94.6 & 86.6 & 79.5 & 55.6 & 93.6 & 96.1 & 79.8 & 80.7 & 89.2 & 89.0 & 86.1 & 57.1 & 82.3 & 77.8 \\
Ours & R18S & 96.0 & 89.5 & 84.3 & 67.2 & 95.9 & 94.2 & 91.0 & 81.5 & 93.8 & 89.9 & 89.1 & 58.2 & 85.9 & 82.8 \\
TENT~\cite{wang2021tent} & R50S & 58.9 & 40.1 & 50.2 & 23.6 & 22.6 & 25.3 & 29.8 & 24.8 & 22.9 & 30.2 & 45.1 & 20.1 & 32.8 & 31.4 \\
Ours & R18S & 90.5 & 65.2 & 79.6 & 38.8 & 26.7 & 12.9 & 51.6 & 59.9 & 44.2 & 46.0 & 71.1 & 24.0 & 50.9 & 49.9 \\
\bottomrule
\end{tabular}}
\vspace{2mm}
\caption{Classification accuracy of our \method on VisDA-C (validation)~\cite{peng2017visda} across all categories and averaged over classes (avg.) and images (acc.). R18/50/101S denotes ResNet-18/50/101 randomly initialized from scratch and R18/50/101P denotes ResNet-18/50/101 pretrained on ImageNet.
}
\label{tab-visda-val}
\vspace{-8.5mm}
\end{table}

\textbf{VisDA train $\rightarrow$ val} Table~\ref{tab-visda-val} compares our method with state-of-the-art unsupervised (upper part) and test-time (lower part) domain adaptation approaches from VisDA-C train to val splits.
The proposed on-target adaptation significantly improves the existing test-time adaptation methods.
It is worth noting that all these existing methods need to keep the same architecture when joint-training or fine-tuning on the source model.
On the contrary, our method could utilize a much more lightweight model, such as ResNet-18 as shown in this table.
For example, our \method brings 18+ points improvement compared to test-time adapted teacher TENT, while reducing over 50\% parameters and runtime flops, and 75\% memory consumption at each feed-forward of target model.
\begin{table}[t]
    \centering
    \resizebox{0.95\textwidth}{!}{
    \Huge
    \begin{tabular}{lrrrrc|clrrrrc|clrrrr}
     \multicolumn{3}{r}{source-only} & \multicolumn{2}{c}{ours} & ~ & ~  & \multicolumn{3}{r}{TENT} & \multicolumn{2}{c}{ours} & ~ & ~ & \multicolumn{3}{r}{SHOT} & \multicolumn{2}{c}{ours} \\
    network & avg. & acc. & avg. & acc. & ~ & ~ & network & avg. & acc. & avg. & acc. & ~ & ~ & network & avg. & acc. & avg. & acc. \\
    \midrule
    R50S & 22.1 & 23.9 & 30.9 & 33.9 & ~ & ~ & R50S & 34.2 & 34.3 & 49.3 & 51.2 & ~ & ~ & R101P & 89.3 & 88.4 & 91.7 & 91.6 \\
    R50P & 34.4 & 37.7 & 39.8 & 43.5 & ~ & ~ & R50P & 60.4 & 62.4 & 74.8 & 77.7 & ~ & ~ & R50P & 76.4 & 78.0 & 81.1 & 83.0 \\
    \bottomrule
    \end{tabular}}
    \vspace{2mm}
    \caption{Classification accuracy of our \method supervised by three teachers: source-only, SHOT~\cite{liang2020we}, and TENT~\cite{wang2021tent} on VisDA-C (test)~\cite{peng2017visda}. R50/101S denotes ResNet-50/101 randomly initialized from scratch and R50/101P denotes ResNet-50/101 pretrained on ImageNet.
    }
    \label{tab-visda-test}
    \vspace{-5mm}
\end{table}
\textbf{VisDA train $\rightarrow$ test} Table~\ref{tab-visda-test} compares our method with state-of-the-art unsupervised (upper part) and test-time (lower part) domain adaptation approaches from VisDA-C train to test splits.
Similar to table~\ref{tab-visda-val}, our method dramatically improves the performance of all kinds of teacher models.
In the following two paragraphs, we discuss the potential application of on-target adaptation \emph{without contrastive learning}.
When test-time data is not sufficient enough to finish the contrastive learning,
we could skip contrastive learning on target data (stage 2) of the proposed \method.
In other words, we directly fine-tune the target model initialized by the source model.
We believe that adaptation performance could be further improved once the contrastive learning could no longer be data-hungry or target domain data could be abundant.

\begin{figure}[t]
	\begin{minipage}[t]{0.51\textwidth}
		\captionsetup{type=table} %
		\centering
\resizebox{0.95\textwidth}{!}{
\begin{tabular}{lccr}
method & network & accuracy \\
\midrule
Anisotropic~\cite{mishra2020learning} & ResNet-50 & 24.5 \\
Debiased~\cite{li2020shape} & ResNet-50 & 28.4 \\
Crop~\cite{hermann2019origins} & ResNet-50 & 30.9 \\
RVT~\cite{mao2021rethinking} & DeiT-B & 36.0 \\
\midrule
TENT~\cite{wang2021tent} & ResNet-50 & 35.6 \\
Ours & ResNet-18 & 37.5 \\
Ours & ResNet-50 & 40.5 \\
\bottomrule 
\end{tabular}}
		\caption{
			Classification accuracy on ImageNet-Sketch~\cite{wang2019learning} benchmark.
		}
		\label{tab-imagenet-sketch}
	\end{minipage}
	\hfill
	\begin{minipage}[t]{0.47\textwidth}
		\captionsetup{type=table} %
		\centering

\resizebox{0.95\textwidth}{!}{
\begin{tabular}{lccr}
method & network & accuracy \\
\midrule
DANN~\cite{ganin2015unsupervised} & ResNet-50 & 57.6 \\ 
DAN~\cite{long2015learning} & ResNet-50 & 56.3 \\
CDAN+E~\cite{long2018conditional} & ResNet-50 & 65.8 \\
CDAN+BSP~\cite{chen2019transferability} & ResNet-50 & 66.3 \\
SAFN~\cite{xu2019larger} & ResNet-50 & 67.3 \\
CDAN+TN~\cite{wang2019transferable} & ResNet-50 & 67.6 \\
\midrule
TENT~\cite{wang2021tent} & ResNet-50 & 53.8 \\
Ours & ResNet-18 & 56.8 \\
\bottomrule 
\end{tabular}}
		\caption{Classification accuracy on Office-Home~\cite{venkateswara2017deep} benchmark.}
		\label{tab-office-home}
	\end{minipage}
	\vspace{-5mm}
\end{figure}

\textbf{ImageNet $\rightarrow$ Sketch}
Table~\ref{tab-imagenet-sketch} reports the empirical results on generalization regarding ImageNet/Sketch as source/target domain. For on-target adaptation, we try two student models: model as same as the teacher model (ResNet-50) and small supervised model pretrained on ImageNet (ResNet-18).
Our \method additionally brings $\sim$5/3\% improvements compared to the teacher model with the same/shallower student models,
where the teacher model, TENT, already outperforms the previous state-of-the-art by $\sim$5\%.

\textbf{Office Home}
Table~\ref{tab-office-home} compares our method with state-of-the-art unsupervised (upper part) and test-time (lower part) domain adaptation approaches for the various domain shifts in Office Home.
Our method advances the average accuracy over all domain shifts of teacher model (TENT) by 3\%.

\subsection{On-target class distribution learning}
\label{sec:exp-long-tailed}

In this section, we argue for calibrating classifier during the test-time, without any modification on training procedure, aiming at long-tailed recognition task.
Here we treat the long-tailed data as the source domain while regarding the class-balanced data as the target domain.
During training, instance-balanced sampling provides a generalizable representation to start with.
Then the classifier is re-scaled during the test-time on the class-balanced data.

First, we train the source domain model with the instance-balanced sampling, which samples each sample with the same probability.
In this way, the learned classifier has a higher prior probability on the head compared to the tail.
Then we calibrate the parameters of the classifier while freezing the feature with test images and pseudo labels, following the practice of our on-target adaptation.
It is worth mentioning that we do not utilize contrastive learning to re-initialize the representation on target data or tune the feature part in teacher-student (stage 3).
The major reason for such a choice is to follow the practice of LWS~\cite{kang2019decoupling}, which points out that the domain shift only exists within class distribution so that only classifier needs to be calibrated.

The empirical results indicate that our method could automatically calibrate the categorical prior and adaptive fit the test data distribution without the access of training data.
Table~\ref{tab-long-tail} demonstrate that our test-time on-target adaptation could achieve comparable performance compared to state-of-the-art training-time methods on ImageNet-LT and iNaturalist-18 datasets.
We report results for two popular ConvNets, ResNet-50 and ResNet-101, as teacher models trained on long-tailed data.
Our method achieves comparable overall performance with the train-time method (LWS) on both datasets.
Compared to the vanilla ResNet-50 and ResNet-101, our fully test-time method significantly improves the overall performance by a large margin.
Considering the accuracy of few-shot (less than 20 samples) categories,
our method outperforms the train-time practice on three out of four cases,
extending the usage scenarios of train-time long-tailed recognition methods.
\se{this is an important outcome of your model but this section is devoted more into the results and less to more insight into why this method could calibrate this well}

\begin{table}[t]
\centering
\resizebox{0.8\textwidth}{!}{
\begin{tabular}{lrrr>{\columncolor[gray]{.9}}rrrr>{\columncolor[gray]{.9}}r}
    & \multicolumn{4}{c}{iNaturalist18} & \multicolumn{4}{c}{ImageNet-LT} \\
method & many & medium & few & acc. & many & medium & few & acc. \\
\midrule
ResNet-50 & 72.2 & 63.0 & 57.2 & 61.7 & 64.0 & 33.8 & 5.8 & 41.6 \\
+ LWS~\cite{kang2019decoupling} & 65.0 & 66.3 & 65.5 & 65.9 & 57.1 & 45.2 & 29.3 & 47.7 \\
+ Ours & 64.2 & 66.3 & 65.9 & 65.9 & 55.7 & 46.0 & 28.6 & 47.4 \\
\midrule
ResNet-101 & 75.9 & 66.0 & 59.9 & 64.6 & 66.6 & 36.8 & 7.1 & 44.2 \\
+ LWS~\cite{kang2019decoupling} & 69.6 & 69.1 & 67.9 & 68.7 & 60.1 & 47.6 & 31.2 & 50.2 \\
+ Ours & 66.5 & 69.1 & 68.3 & 68.5 & 58.9 & 48.7 & 31.8 & 50.3 \\
\bottomrule 
\end{tabular}}
\vspace{2mm}
\caption{
    Comparing our \method performance with learnable weight scaling (LWS)~\cite{kang2019decoupling} on long-tailed benchmarks including iNaturalist18~\cite{van2018inaturalist} and ImageNet-LT~\cite{liu2019large}. Note that LWS works during training while our \method only need to calibrate classifier during testing. 
}
\label{tab-long-tail}
\vspace{-5mm}
\end{table}

\subsection{Ablation Studies}
\label{sec:exp-ablation}

\textbf{Stage 0: network \& initialization}
\begin{table}[t]
\centering
\resizebox{0.95\textwidth}{!}{
\Huge
\begin{tabular}{lccrrrrc|clccrrrr}
& imagenet & source & \multicolumn{2}{c}{SHOT} & \multicolumn{2}{c}{ours} & ~ & ~ & & imagenet & source & \multicolumn{2}{c}{TENT} & \multicolumn{2}{c}{ours} \\
network & pretrain & only & avg. & acc. & avg. & acc. & ~ & ~ & network & pretrain & only & avg. & acc. & avg. & acc. \\
\midrule
ResNet-50 & \xmark & \xmark &  49.7 & 48.3 & 69.1 & 67.3 & ~ & ~ & ResNet-50 & \xmark & \xmark & 32.8 & 31.4 & 50.9 & 49.9 \\
ResNet-50 & \cmark & \xmark & 75.0 & 74.5 & 77.8 & 78.6 & ~ & ~ & ResNet-50 & \cmark & \xmark & 60.9 & 60.2 & 75.1 & 73.8 \\
ResNet-101 & \cmark & \xmark & 82.3 & 77.8 & 85.9 & 82.8 & ~ & ~ & ResNet-18 & \xmark & \xmark & 34.0 & 33.1 & 51.4 & 51.4 \\
\midrule
ResNet-101 & \cmark & \cmark & 49.9 & 55.5 & 60.0 & 65.6 & ~ & ~ & ResNet-50 & \xmark & \cmark & 17.8 & 21.8 & 22.3 & 29.1 \\
\bottomrule 
\end{tabular}}
\vspace{2mm}
\caption{
    Classification accuracy on VisDA-C (validation)~\cite{peng2017visda}. 
    ``Imagenet pretrain'' indicates whether we utilize ResNet pretrained on ImageNet at stage 0. 
    ``Source only'' indicates whether we skip test-time adaptation (stage 1) and directly use the source model to generate pseudo labels at stage 3.
}
\label{tab-teacher-arch}
\vspace{-7mm}
\end{table}
The upper part of Table~\ref{tab-teacher-arch} presents the numbers of SHOT/TENT with ResNet in various depths (18, 50, 101) and initializations (from scratch, ImageNet pretrain).
We observe that our \method consistently improves the teacher models with various model architectures and initializations, which indicates the usability and versatility of the proposed framework.
When adapting from synthetic to real-world domains, the ImageNet pretrained model should not be utilized to start with, due to the learned inductive bias of its parameters.
Therefore we also experiment on training from scratch for teacher models.
Larger capacity does not lead to better generalization when training from scratch.
On the contrary, We observe that a deeper ImageNet pretrained ConvNet provides a stronger inductive bias from the beginning.

\textbf{Stage 1: test-time adaptation}
The lower part of Table~\ref{tab-teacher-arch}
presents the numbers of source-only models as teacher,
without any test-time adaptation (TENT/SHOT).
The final accuracy after teacher-student suffers from the poorer quality of the initial pseudo label.
ImageNet pretraining could alleviate such a phenomenon, but the numbers with test-time adapted teachers are still significantly better than the source-only ones.
Comparing these empirical results with table~\ref{tab-visda-val}, TENT boosts the initial/final accuracy by 15.0/28.6 points, while SHOT brings 32.4/25.9 points improvement.
In a word, test-time adaptation should be leveraged in preparation for trustworthy pseudo labels.

\begin{figure}[t]
	\begin{minipage}{0.64\textwidth}
		\captionsetup{type=table} %
		\centering
    \centering
    \resizebox{0.95\textwidth}{!}{
    \Huge
    \begin{tabular}{lrrc|clrr}
    method & avg. & acc. & ~ & ~ & method & avg. & acc. \\
    \midrule
    TENT & 32.8 & 31.4 & ~ & ~ & Ours (MoCo) & 50.9 & 49.9 \\
    \midrule
    VisDA-C train & 45.3 & 43.3 & ~ & ~ & SwAV & 50.5 & 49.4 \\
    ImageNet & 47.3 & 46.2 & ~ & ~ & SimSiam & 48.7 & 47.6 \\
    ImageNet (MoCo) & 46.6 & 45.5 & ~ & ~ & Barlow Twins & 46.3 & 44.8 \\
    \bottomrule
    \end{tabular}}
		\caption{Classification accuracy on VisDA-C (validation)~\cite{peng2017visda}. Left side: Ablation results on the student model with various initialization. Right side: Ablation results on the contrastive learning method using MoCo~\cite{chen2020improved}, SwAV~\cite{caron2020unsupervised}, SimSiam~\cite{chen2020exploring}, and Barlow Twins~\cite{zbontar2021barlow}.
		}
		\label{tab-student-init}
	\end{minipage}
	\hfill
	\begin{minipage}{0.33\textwidth}
		\centering
		\includegraphics[width=\linewidth]{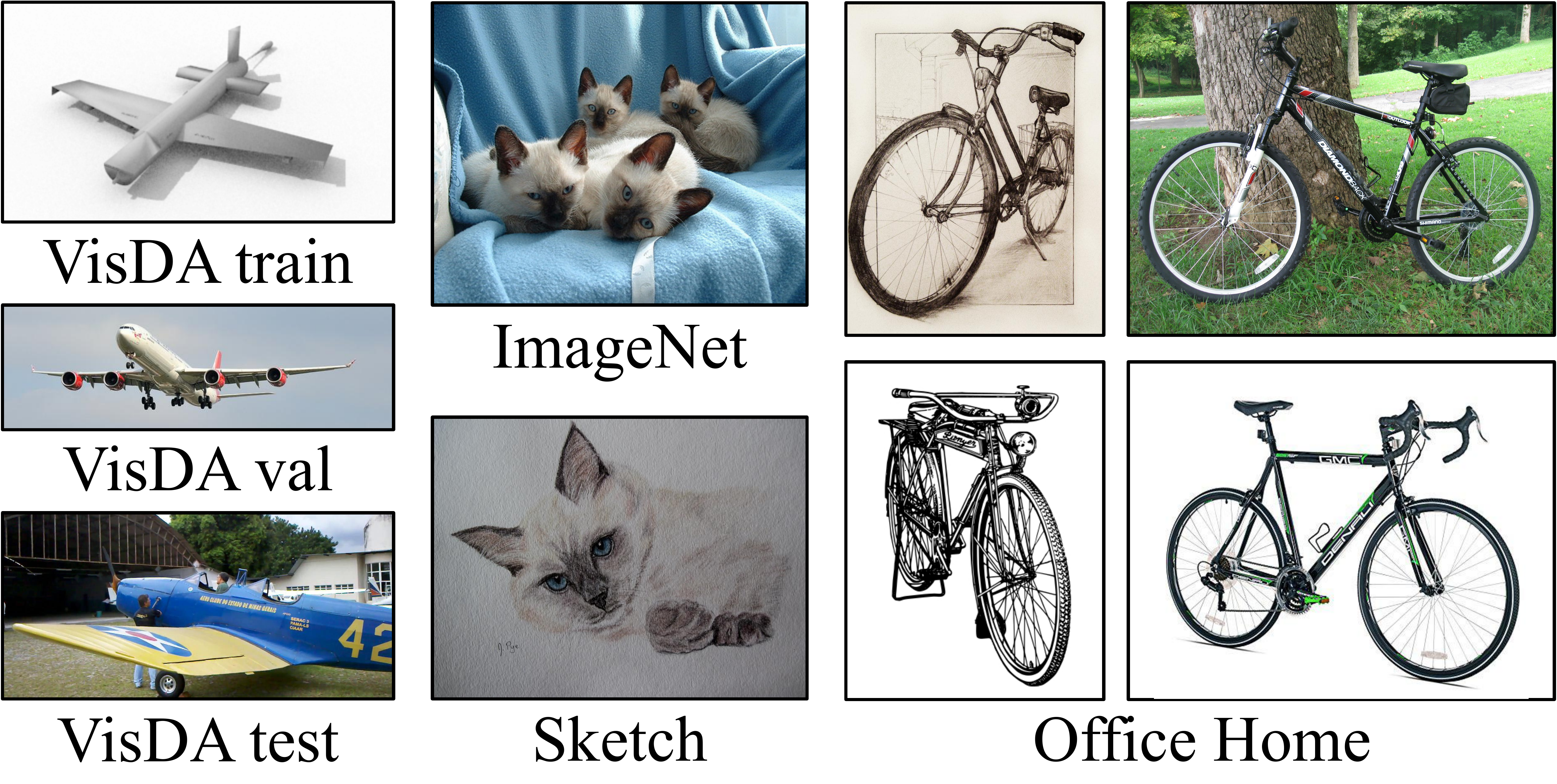}
		\caption{
			Example images from VisDA-C~\cite{peng2017visda}, ImageNet~\cite{russakovsky2015imagenet}, ImageNet Sketch~\cite{wang2019learning}, and Office-Home~\cite{venkateswara2017deep}.
		}
		\label{fig-domain-shift}
	\end{minipage}
	\vspace{-1mm}
\end{figure}

\textbf{Stage 2: on-target feature}
The left part of Table~\ref{tab-student-init} presents the ablation study of student architecture and initialization.
When the student model is not initialized on target data, such as source data (VisDA-C train) or the external large dataset (ImageNet),
the overall accuracy drops 4+ points, which indicates the necessity of on-target feature learning.
We also ablate the same contrastive learning algorithm (MoCo v2) on a different data source, such as VisDA-C val and ImageNet.
The empirical results indicate that more external data does not bring any advantage, which echos our statement on the target-specific representation learning.

\textbf{Stage 2: contrastive learning}
The right part of Table~\ref{tab-student-init} presents the results with various contrastive learning frameworks, including SwAV, SimSiam, Barlow Twins.
We train these contrastive learned models with the same number of epochs to have a fair comparison.
Compared to the performance of the teacher model, all these learned features bring a noticeable improvement, achieving the comparable numbers with the reference performance of MoCo v2.
We observe that our \method is not sensitive to the choice of contrastive learning method.
In this way, our on-target adaptation could be further improved by introducing a more advanced contrastive learning approach in the future.

\begin{table}[t]
\centering
\resizebox{0.95\textwidth}{!}{
\Huge
\begin{tabular}{ccrrr>{\columncolor[gray]{.9}}rrrrrrrrr}
teacher  & soft & phase 0 & phase 1 & phase 2 & phase 3 & phase 4 & phase 5 & phase 6 & phase 7 & phase 8 & phase 9 \\
\midrule
\multirow{2}{*}{TENT} & \xmark & 32.8 & 44.2 & 48.2 & 50.9 & 52.7 & 54.3 & 56.0 & 57.0  & 58.1 & 58.9 \\
 & \cmark & 32.8 & 44.3 & 53.5 & 56.4 & 59.9 & 61.4 & 63.6 & 64.2 & 65.1 & 65.6 \\
\midrule
\multirow{2}{*}{SHOT} & \xmark & 82.3 & 84.8 & 85.5 & 85.9 & 86.2 & 86.3 & 86.3 & 86.3 & 86.2 & 86.3 \\
 & \cmark & 82.3 & 84.7 & 85.2 & 85.6 & 85.2 & 85.7 & 85.0 & 85.4 & 84.9 & 85.2 \\
\bottomrule 
\end{tabular}}
\vspace{2mm}
\caption{Classification accuracy of our \method on VisDA-C (validation)~\cite{peng2017visda}. ``Soft'' indicates whether we replace cross entropy loss (hard label) with Kullback–Leibler divergence (soft label) for the even number of phases. Note that our default number of phases (phase 3) is highlighted.
}
\label{tab-interaction-phase}
\vspace{-5mm}
\end{table}

\textbf{Stage 3: more phases}
Table~\ref{tab-interaction-phase} presents the detailed numbers for each phase during teacher-student.
We observe that the first phase already significantly outperforms the test-time adapted teacher model, which is also the state-of-the-art practice.
The following several phases gradually improve the results, taking the last generation student as the next teacher.
Considering the speed-accuracy trade-off, we choose to have three phases as our default setup, even though more phases could lead to a better result.
For example, 9-phase (3$\times$) optimization brings up around 9 points improvement compared to our default 3-phase (1$\times$) one with TENT as the initial teacher model.

\textbf{Stage 3: soft label}
Table~\ref{tab-interaction-phase} presents the ablation study on the design choice of loss function.
Our default setup only utilizes hard labels with cross-entropy loss.
Actually, our framework also benefits from the soft label with Kullback–Leibler divergence loss, following the popular practice of knowledge distillation~\cite{hinton2015distilling}.
We observe that the mix of both hard and soft label bring up the best performance.
we replace the cross-entropy loss (hard label) with Kullback–Leibler divergence (soft label) for the \emph{even} number of phases.
We set the number of epochs as one for all these interpolated soft label phases for a more computational-friendly practice.
The known drawback is that the soft label part typically needs specific tuning on learning rate, loss weight, temperature, and so on.
Existing works~\cite{berthelot2019mixmatch,berthelot2019remixmatch,sohn2020fixmatch} discuss sharpening (temperature) and thresholding (confidence threshold) to improve the performance of semi-supervised learning.
Instead, we only ablate the default Kullback–Leibler divergence loss without bells and whistles like temperature and confidence threshold.
Our default training objective chooses to be the most robust hard label with cross-entropy criterion for all the other experiments.

\section{Conclusion}
In fully test-time adaptation, a model needs to adaptively update its parameters to deal with unseen scenarios, where only trained source-domain model and unlabeled target-domain data are given.
The proposed on-target framework significantly improves the generalization performance for both domain adaptation and long-tailed recognition. %
The performance gain comes from the following three stages: 1) test-time adaptation; 2) contrastive learning; 3) teacher-student.
We first deploy the test-time adapted source model as a teacher network to generate pseudo-labels. %
Then we leverage contrastive learning to initialize target-feature on unlabeled target data. %
At last, we propose a simple yet effective teacher-student framework to transfer knowledge from teacher to student.
Extensive empirical studies demonstrate that our method achieves results surpassing or comparable to the state-of-the-art methods, indicating the generalizable effectiveness on various visual recognition tasks.

\clearpage
\newpage
{
\begin{spacing}{0.96}
\bibliographystyle{iclr2021_conference}
\bibliography{ref}
\end{spacing}
}

\end{document}